\documentclass[conference,a4paper]{IEEEtran}
\IEEEoverridecommandlockouts

\usepackage{cite}
\usepackage{amsmath,amssymb,amsfonts}
\usepackage{algorithmic}
\usepackage{graphicx}
\usepackage{textcomp}
\usepackage{xcolor}
\def\BibTeX{{\rm B\kern-.05em{\sc i\kern-.025em b}\kern-.08em
    T\kern-.1667em\lower.7ex\hbox{E}\kern-.125emX}}
\begin{document}

\title{Semantic-Aware Predictive Mapping for Exploration and Navigation}

\author{
\IEEEauthorblockN{1\textsuperscript{st} Kenneth J.K. Ong}
\IEEEauthorblockA{
\textit{NEAR Lab} \\
\textit{AI.R STC, GTO, ST Engineering}\\
jookiat.kenneth.ong@stengg.com}
\and
\IEEEauthorblockN{2\textsuperscript{nd} William W.J. Teo}
\IEEEauthorblockA{
\textit{NEAR Lab} \\
\textit{AI.R STC, GTO, ST Engineering}\\
\textit{Dept. of Mechanical Engineering, CDE} \\
\textit{National University of Singapore}\\
weijie.william.teo@stengg.com}
}

\IEEEpubid{%
\makebox[\columnwidth][l]{%
\parbox[t]{0.98\columnwidth}{%
\raggedright
\footnotesize
\textcopyright~2026 IEEE. Personal use of this material is permitted.
Permission from IEEE must be obtained for all other uses.
}}%
\hspace{\columnsep}%
\makebox[\columnwidth]{}%
}

\maketitle

\begin{abstract}
Predictive mapping can support robotic exploration and navigation by estimating unseen geometric layouts from partial occupancy observations. However, occupancy-only representations may fail to distinguish semantically different structures with similar geometry. This is particularly relevant for indoor doors, which may appear as occupied cells like walls but indicate possible connected rooms or corridors beyond the observed region. This work investigates whether semantic door cues improve predictive geometric occupancy mapping around such ambiguous regions. We modify a subset of the CogniPlan dataset by inserting door-induced ambiguities into partial occupancy maps while keeping the ground-truth layouts unchanged. We compare a geometry-only control model with a semantic-cued model trained on the same modified dataset, where the semantic-cued model receives an additional door channel. Evaluation uses L1 error, F1 score, and Intersection over Union (IoU) over both the full map and a 10-pixel door-region mask. Full-map performance remains broadly similar between models, but localized door-region results show a clear qualitative improvement: L1 decreases from 0.004342 to 0.000025, while F1 and IoU improve from 0.031311 and 0.015905 to 1.000000 and 1.000000, respectively. These results suggest that semantic cues can improve predictive occupancy completion in regions where geometric observations alone are ambiguous.
\end{abstract}


\begin{figure}[t]
\centering{\includegraphics[width=0.3\textwidth]{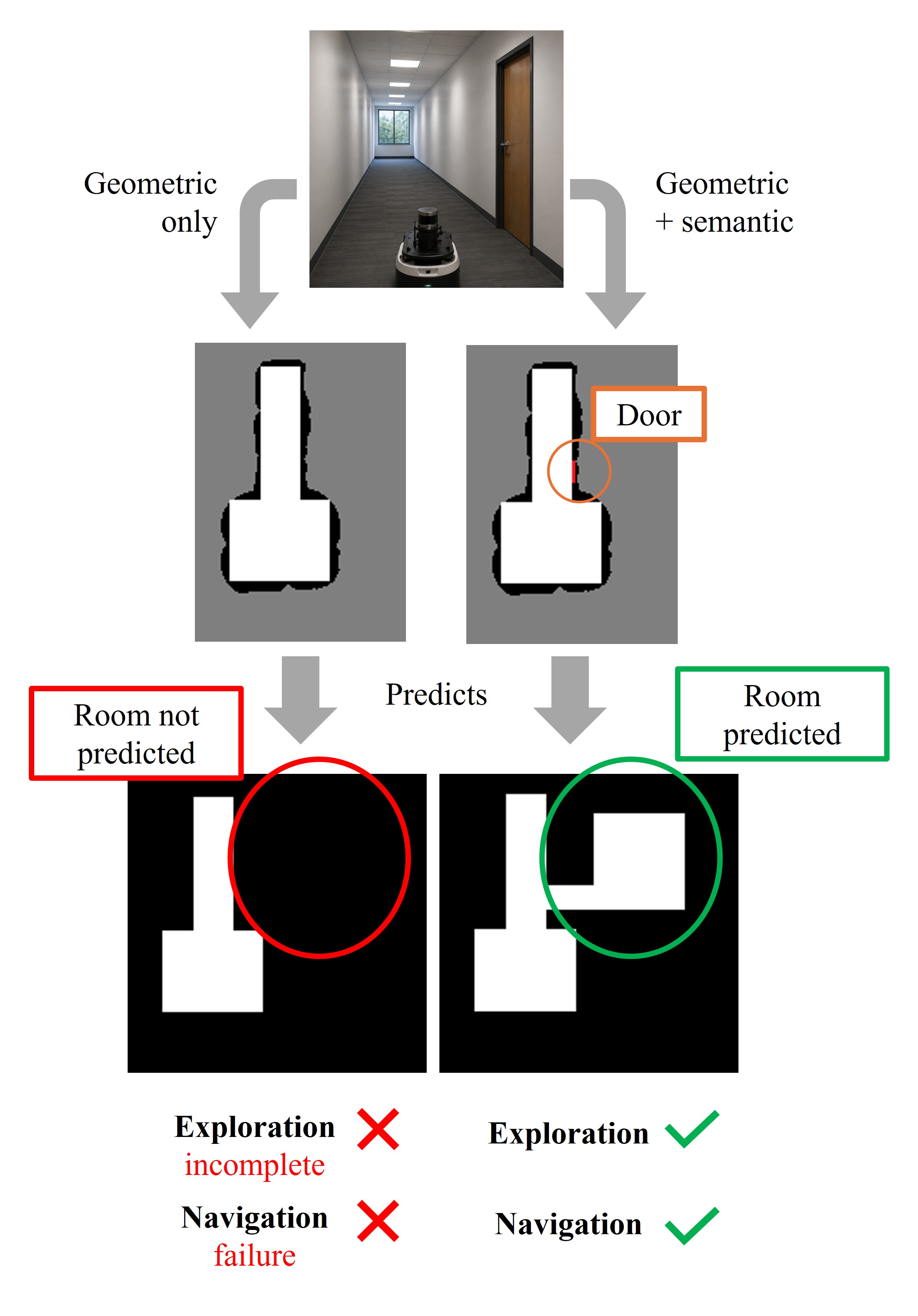}}
\caption{Comparison between purely geometric predictive mapping and semantic-aware predictive mapping in indoor environments containing doors. When doors are represented identically to walls in occupancy maps, geometry-only prediction fails to infer connected unseen spaces beyond the door, resulting in incomplete room prediction and reduced exploration and navigation effectiveness. By incorporating semantic cues indicating the presence of doors, the predictive model is able to infer plausible connected structures beyond geometrically occluded regions, improving the completeness of predicted layouts. The corridor image shown is generated using OpenAI DALL·E for illustrative purposes.}
\label{fig1}
\end{figure}

\section{Introduction}
Autonomous exploration and navigation in unknown environments remain fundamental challenges in robotics. Mobile robots operating in partially observed environments commonly rely on geometric representations such as occupancy maps constructed from onboard sensors, including LiDAR and RGB-D cameras. Exploration and navigation are then typically performed using geometric planning strategies, including frontier-based exploration~\cite{yamauchi1997frontier}, next-best-view methods~\cite{bircher2016receding,zhu2021dsvp}, learning-based methods~\cite{zhu2018deep,li2019deep,kamalova2022occupancy,tao2024learning}, sampling-based planning~\cite{kuffner2000rrt,gammell2015batch,karaman2011sampling}, and classical path planning~\cite{hart1968formal,koenig2002d,likhachev2005anytime}. Despite differences in planning strategy, these methods primarily reason over geometric occupancy structure and traversability.

Predictive mapping approaches~\cite{tao2024learning,8793769,luperto2020exploration,georgakis2022uncertainty,ho2025mapex,wei2021occupancy,baek2025pipe,katyal2019uncertainty,wang2025cogniplan} extend this setting by inferring unseen geometric layouts from partial observations. For example, CogniPlan~\cite{wang2025cogniplan} predicts and combines multiple possible map layouts to support uncertainty-aware waypoint planning in partially observed environments. In parallel, semantic prediction and semantic mapping methods have been explored for reasoning about unobserved spaces and supporting navigation, including semantic annotations over BEV maps~\cite{zhang2025mapnav,roddick2020predicting}, occupancy maps~\cite{hiller2019learning}, and point clouds~\cite{yang2017semantic,pan2025point,wang2019associatively}. For instance, Zhang et al.~\cite{zhang2024imagine} predict the semantic composition of unseen regions, including likely object locations.

\IEEEpubidadjcol

However, geometric predictive mapping and semantic prediction are often treated as separate problems. Predictive mapping methods mainly focus on occupancy completion and traversability estimation, while semantic prediction methods focus on object or scene semantics in unseen regions. As a result, semantic information is rarely used to directly improve predictive geometric layout modeling and spatial connectivity inference.

This limitation is particularly relevant in indoor environments containing doors. From a purely geometric occupancy perspective, closed doors may be represented as occupied obstacles, making them difficult to distinguish from walls. Semantically, however, doors indicate possible room connectivity and the existence of traversable space beyond the current observation boundary. Inferring such hidden connectivity requires reasoning beyond local occupancy geometry alone. Fig.~\ref{fig1} illustrates the motivation for distinguishing geometric occupancy information from semantic door cues in predictive mapping.

In this work, we investigate semantic-aware predictive mapping for indoor exploration and navigation. Specifically, we study whether semantic cues related to doors and room connectivity can improve the prediction of unseen geometric structures behind geometrically occluded regions. While doors are used as the case study, this work highlights the broader role of semantic information in predictive geometric mapping for robotic exploration and navigation.

The main contributions of this work are as follows:
\begin{itemize}
    \item Using doors as a case study, we demonstrate that incorporating semantic information into predictive occupancy mapping can improve the completeness of inferred unseen structures beyond geometrically occluded regions.
    \item We motivate the broader integration of semantic understanding into future predictive world models for robotic exploration and navigation.
\end{itemize}

\section{Methodology}
\subsection{Overview}
This work adopts and extends the predictive mapping framework introduced in CogniPlan. CogniPlan learns to predict unseen geometric layouts from partially observed occupancy maps, enabling improved waypoint planning for exploration and navigation in unknown environments. The original framework primarily relies on geometric occupancy information without explicitly incorporating semantic understanding of the environment.

Our work investigates whether semantic cues can improve predictive geometric reasoning in indoor environments containing semantically meaningful structures such as doors. This work focuses on incorporating semantic information into the predictive mapping pipeline and evaluating its effect on layout prediction quality.

\subsection{Dataset Modification}
To evaluate whether semantic cues can improve predictive geometric mapping around door-induced ambiguity, we modify a subset of the original CogniPlan occupancy-map dataset. The modification is applied only to the partial observation inputs, while the corresponding ground-truth occupancy layouts are kept unchanged. This ensures that all models are evaluated against the same underlying map structure.

In the original occupancy representation, occupied cells do not distinguish between different types of barriers. A wall, closed door, or other occupied structure is represented using the same occupied-space encoding. Consequently, a model that receives only geometric occupancy information cannot determine whether an occupied segment corresponds to a permanent wall or a door that may imply connected traversable space beyond it.

\begin{figure}[t]
    \centering
    \includegraphics[width=0.55\linewidth]{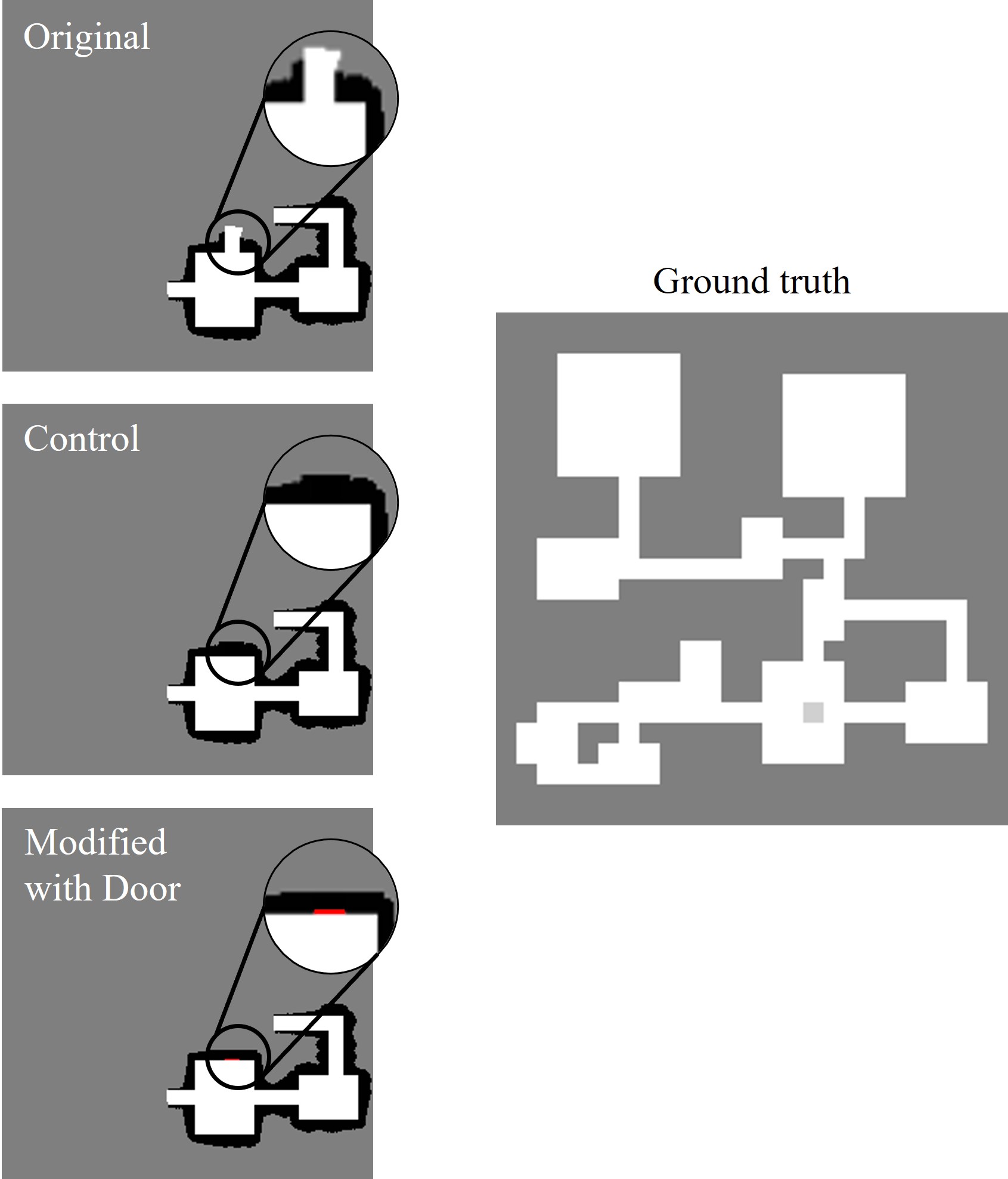}
    \caption{Example of the dataset modification procedure. The original partial occupancy input is modified by inserting door segments at selected room or corridor entrances. In the geometry-only modified input, inserted doors are encoded as occupied cells and are therefore indistinguishable from walls. In the semantic-aware input, the same door regions are additionally marked using a semantic door cue, shown in red for visualization. The ground-truth occupancy layout is unchanged across all variants.}
    \label{dataset_modification}
    \vspace{-3pt}
\end{figure}

To introduce this ambiguity, we insert artificial door segments at selected room entrances and corridor entrances in the partial occupancy maps as illustrated in Fig.~\ref{dataset_modification}. In the geometry-only modified input, each inserted door segment is encoded as occupied space, identical to surrounding walls. This produces a door-induced ambiguity. The input suggests that the entrance is blocked, while the unchanged ground-truth map still contains the connected room or corridor beyond that location.

For training, we use maps from 500 rooms in the original CogniPlan dataset, of which 50 rooms are modified to include artificial door segments. Each room comprises multiple layouts, with each individual layout containing one to two doors. Evaluation is conducted on a separate, dedicated validation set of 5 modified rooms containing door-induced ambiguity, specifically spanning 7 distinct layouts and 8 total doors.

The key property of the modified dataset is that the ground-truth occupancy map remains unchanged across the geometry-only and semantic-aware variants. Only the partial observation input is modified. This allows the experiment to test whether semantic information about door locations helps the model predict connected unseen regions that are geometrically occluded in the partial observation.

\subsection{Experimental Setup and Evaluation}
\begin{figure}[t]
\centering
\includegraphics[width=0.65\linewidth]{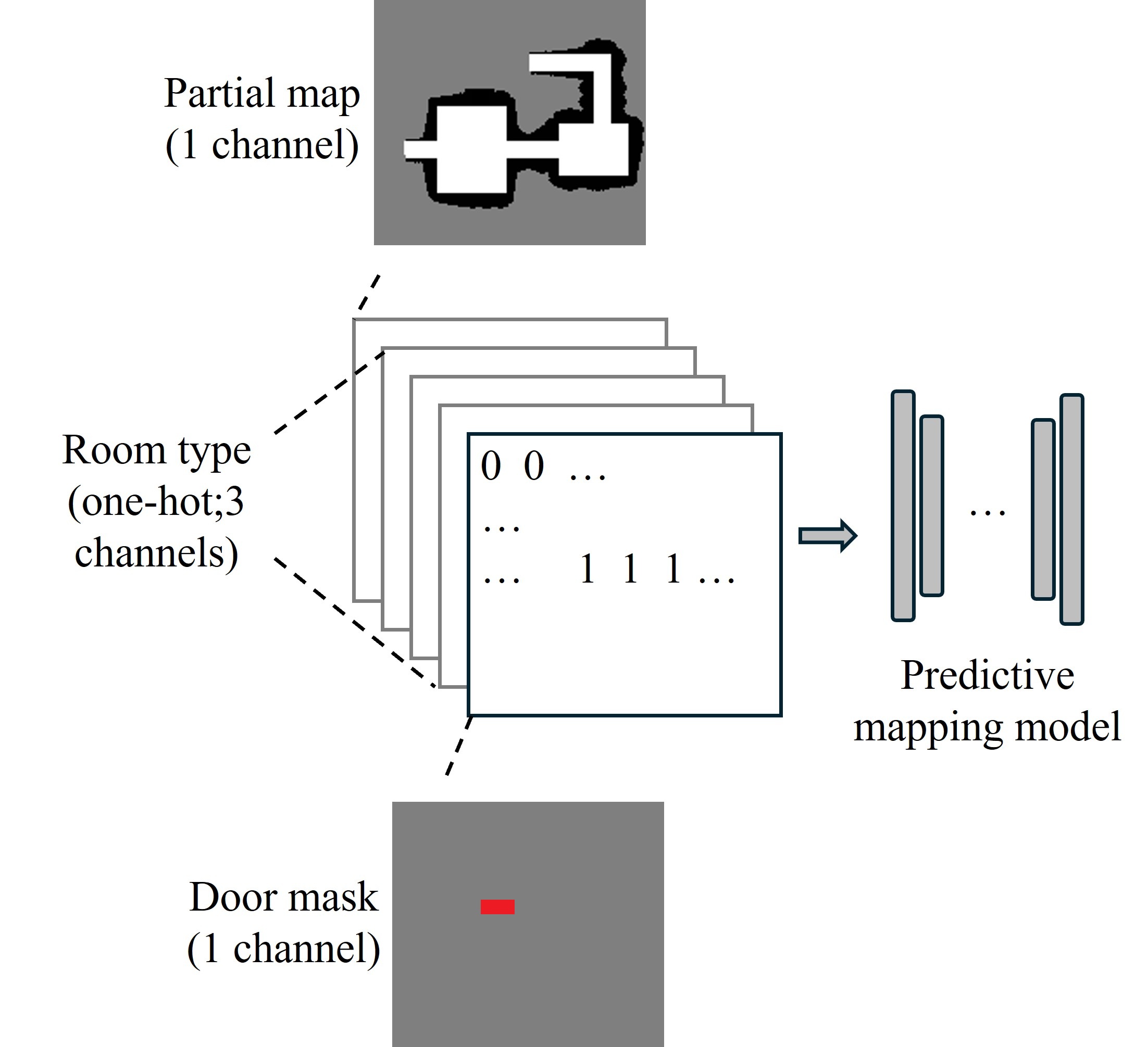}
\caption{Illustration of the semantic-aware input construction. Door regions are shown as red pixels in the modified occupancy map. These red door regions are converted into an additional semantic door channel and concatenated with the original occupancy input before being passed into the predictive mapping model. The ground-truth occupancy layout remains unchanged.}
\label{dataset_modification2}
\end{figure}

Two predictive mapping configurations are evaluated in this work:

\begin{itemize}
    \item \textbf{Geometry-Only Control Model} -- A CogniPlan-based model trained on the door-modified dataset without the semantic door channel. In this setting, door regions are encoded only as occupied cells, identical to walls. This model serves as the control condition and evaluates whether a model trained on the modified dataset can infer hidden connectivity using geometric occupancy information alone.
    \item \textbf{Semantic-Cued Predictive Model} -- A CogniPlan-based predictive mapping model trained on the same door-modified dataset, but with an additional semantic door channel. The original occupancy input is preserved, while door regions are provided as explicit semantic cues as illustrated in Fig.~\ref{dataset_modification2}. This model receives semantic information distinguishing doors from structural walls.
\end{itemize}

Both models are trained and evaluated on the same modified dataset. The only difference between the two configurations is the presence or absence of the semantic door channel. This provides a controlled comparison for evaluating whether explicit semantic door cues improve predictive occupancy mapping around door-induced ambiguous regions.

Both models are trained for 500,000 iterations with a batch size of 32, following the training configuration used in the CogniPlan framework.

For evaluation, we report L1 error, Intersection over Union (IoU), and F1 score, following the predictive mapping metrics used in CogniPlan. These metrics are computed at two levels. All metrics are derived from a single prediction using solely a room-type conditioning vector~\cite{wang2025cogniplan}. To assess overall performance, we gather all pixels from all 5 evaluation rooms into a single combined set and compute the metrics globally once.

First, we report metrics over the full predicted occupancy map to provide an overall measure of predictive mapping performance. Second, since the main focus of this work is whether semantic door cues improve the prediction of connected traversable space beyond doors, we additionally compute localized metrics over the unknown-region cells within a 10-pixel radius of annotated door locations. We estimate 10 pixels to be the typical door-adjacent room boundaries for this dataset. For threshold-dependent metrics (IoU and F1), the outputs—which range from -1 to 1—are binarized using a threshold value of 0.

\begin{figure}[t]
\centering
\includegraphics[width=0.35\linewidth]{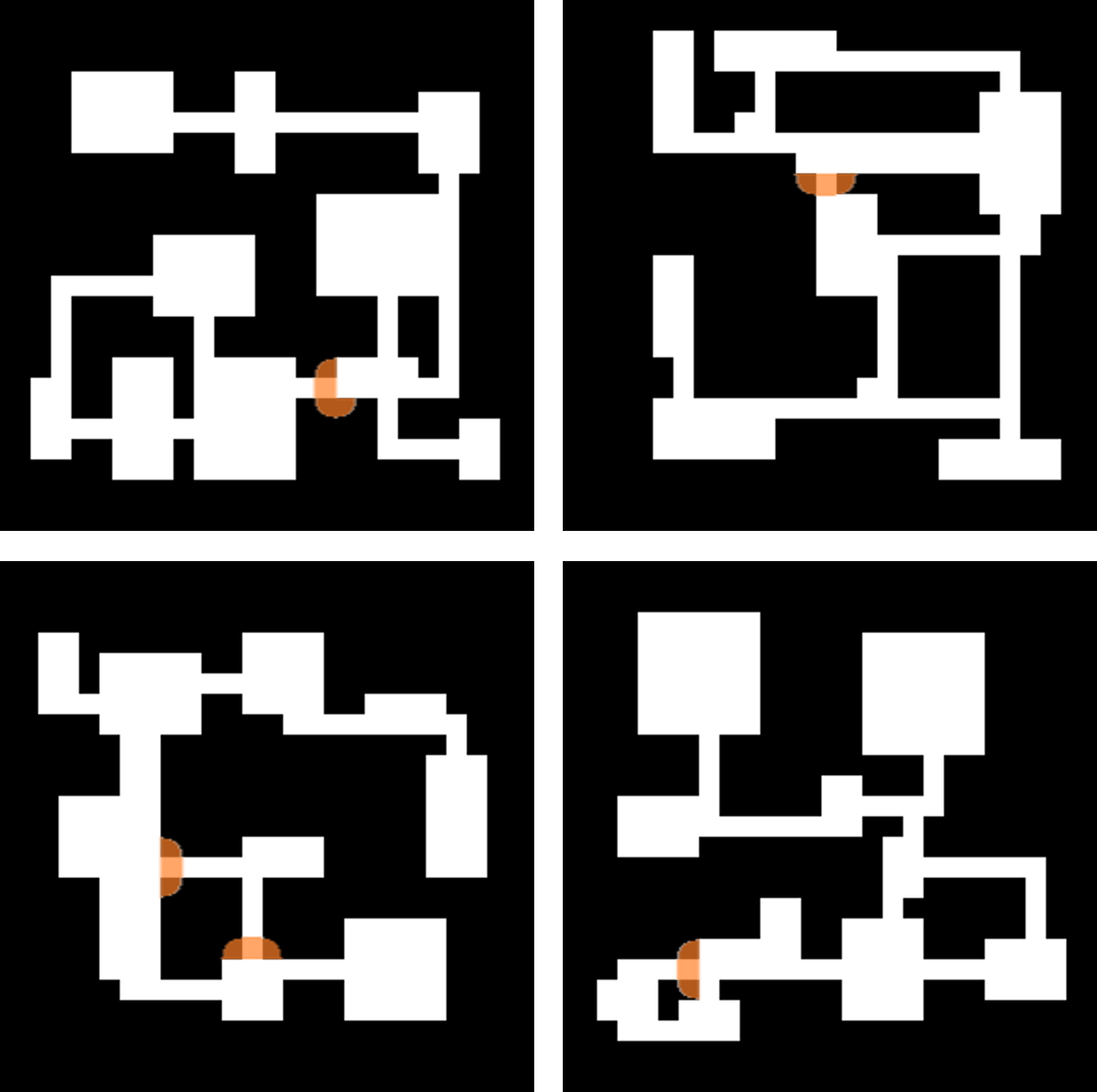}
\caption{Illustration of the localized evaluation region used for door-adjacent quantitative analysis. The highlighted orange area indicates the unknown-region cells within a 10-pixel radius of annotated door locations. Metrics computed over this region evaluate whether the predicted occupancy map captures local connectivity around doors, rather than global map completion alone.}
\label{evaluation_region}
\end{figure}

The localized 10-pixel evaluation region is illustrated in Fig.~\ref{evaluation_region}. This localized evaluation directly targets the regions where door-induced geometric ambiguity occurs, and therefore provides a more focused quantitative measure of how well each model predicts occupancy structure around potential connectivity points.

\section{Results and Discussion}

\begin{table}[t]
\centering
\caption{Quantitative comparison between the geometry-only control model and the semantic-aware predictive model. Metrics are reported over the full map and over unknown cells within a 10-pixel radius around door regions.}
\label{tab:results}
\begin{tabular}{llccc}
\hline
Region & Model & L1 & F1 & IoU \\
\hline
Full Map & Geometric-Only &  0.389160 & 0.681828 & 0.517253 \\
Full Map & Semantic-Cued & 0.388029 & 0.675030 & 0.509469  \\
Door Region & Geometric-Only & 0.004342 & 0.031311 & 0.015905 \\
Door Region & Semantic-Cued & 0.000025 & 1.000000 & 1.000000 \\
\hline
\end{tabular}
\end{table}

Table~\ref{tab:results} reports the quantitative comparison between the geometry-only control model and the semantic-cued predictive model. Metrics are reported at two levels: the full predicted map and the localized door-region mask consisting of unknown cells within a 10-pixel radius around annotated door locations. The full-map metrics provide a measure of overall predictive mapping performance, while the door-region metrics focus on the specific regions where door-induced geometric ambiguity affects connectivity prediction.

For the full-map evaluation, both models obtain broadly similar performance. The semantic-cued model slightly reduces L1 error from 0.389160 to 0.388029, corresponding to an approximately 0.29\% reduction. However, the F1 score decreases from 0.681828 to 0.675030, and the IoU decreases from 0.517253 to 0.509469. These differences are small, suggesting that the semantic door cue does not substantially change global occupancy prediction over the entire map. This is expected, since the semantic modification is localized around door regions, while full-map metrics are dominated by broader layout structures that are not directly affected by door semantics.

The effect of semantic cues is more evident in the localized door-region evaluation. In this region, the semantic-cued model reduces L1 error from 0.004342 to 0.000025, while F1 and IoU increase from 0.031311 and 0.015905 to 1.000000 and 1.000000, respectively. The near-zero performance of the geometry-only control model in the door-region metrics is expected, since doors are encoded identically to walls and therefore provide no indication that connected traversable space may exist beyond them. The small nonzero F1 and IoU values for the geometry-only model are due to partial overlap between some 10-pixel door-region masks and nearby normally predicted regions, rather than successful prediction of behind-door connectivity.

After thresholding, the semantic-cued model agrees perfectly with the ground truth in the door-adjacent unknown regions. However, the door-region mask covers few cells, and the geometry-only model predicts occupied space throughout. These metrics indicate whether the model predicts connectivity beyond doors at all, not the magnitude of a performance gap between models. As confirmed by the full-map metrics, the perfect local agreement does not imply perfect global prediction. It does suggest that the semantic-cued model correctly predicts connected space through door regions in the modified dataset.

\begin{figure}[t]
\centering{\includegraphics[width=0.7\linewidth]{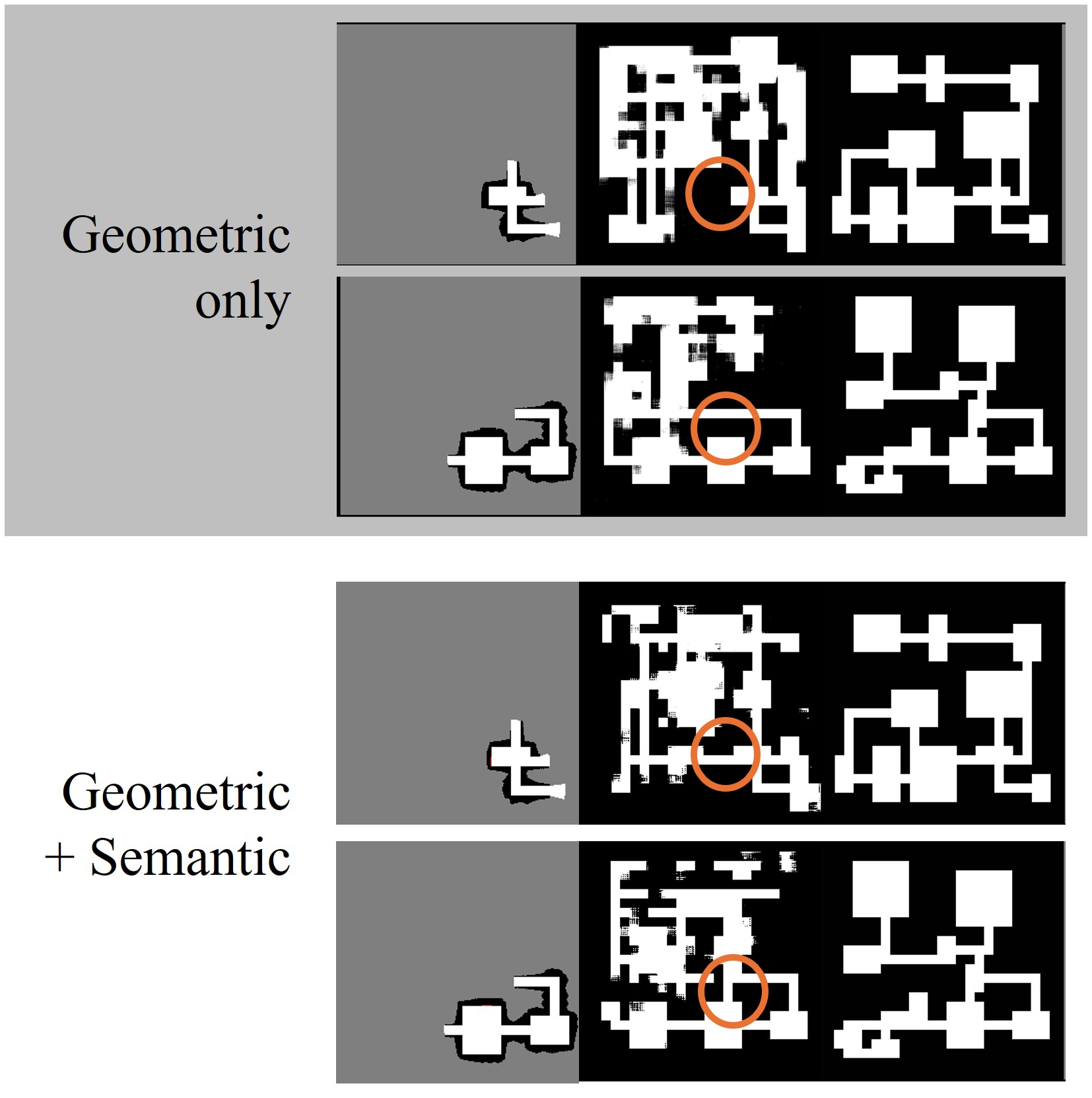}}
\caption{Qualitative comparison between geometry-only and semantic-cued predictive mapping. Each example shows the partial occupancy observation, predicted occupancy layout, and ground-truth map. Orange circles indicate regions where rooms or corridors should exist beyond closed doors. Red pixels indicate semantic door cues in the semantic-cued input.}
\label{res}
\end{figure}

Qualitative comparisons in Fig.~\ref{res} further support this interpretation. In the geometry-only setting, predicted layouts often terminate at door boundaries because doors are represented identically to walls. As a result, rooms or corridors beyond closed-door regions are frequently not predicted. In the semantic-cued setting, the model more consistently predicts connected structures beyond door regions, suggesting that the additional semantic channel helps resolve the ambiguity between walls and doors.

Several recurring behaviors are observed in the geometry-only control model:
\begin{itemize}
    \item Entire rooms located behind doors are frequently omitted from predicted layouts.
    \item Predicted corridors and traversable pathways are often prematurely terminated at door locations.
    \item Potential shortcut connections between rooms and corridors are not inferred, resulting in incomplete navigation topology.
\end{itemize}

These behaviors may negatively affect downstream exploration and navigation tasks. Missing room predictions can reduce exploration completeness, while failure to infer hidden shortcut connections may lead to less efficient path planning. Importantly, the exact shape, size, or layout of the predicted room behind a door does not need to be perfectly accurate for the prediction to be useful. Correctly inferring the existence of connected traversable space is itself valuable, as it indicates that further exploration beyond the door may be worthwhile.

Overall, the quantitative and qualitative results show that semantic door cues have limited effect on global map prediction, but a strong effect on localized connectivity prediction around door-induced ambiguous regions. This supports the use of semantic information as a targeted cue for improving predictive geometric mapping in semantically meaningful regions of indoor environments.

\section{Limitations}
This work has several limitations. First, the dataset is based on a modified subset of the CogniPlan dataset, where door regions are artificially inserted into occupancy-map observations. While this provides a controlled setting for studying door-induced ambiguity, it does not fully capture the variability of real-world sensory observations or indoor layouts.

Second, the semantic information is limited to manually specified door cues. Other semantic structures that may affect navigation and connectivity, such as movable objects, stairs, furniture, and room functions, are not considered. The results should therefore be interpreted as evidence for the usefulness of door semantics, rather than as a complete semantic predictive mapping framework.

Third, although the semantic-cued model shows strong improvement in the localized 10-pixel door-region evaluation, full-map metrics remain broadly similar to the geometry-only control model. This indicates that the semantic cue mainly improves prediction in the targeted door-adjacent regions, rather than global map prediction.

The evaluation set contains 5 rooms, 7 layouts, and 8 total doors. We do not perform significance testing. The perfect door-region F1 and IoU reflect this small sample and should not be taken as precise performance measurements.

Finally, this work evaluates predictive occupancy quality rather than downstream exploration or navigation performance. While the addition of an input channel for door information introduced no observable inference-time overhead, its broader effects on uncertainty aggregation, waypoint selection, exploration coverage, path efficiency, and real-time deployment remain to be evaluated.

\section{Conclusion and Future Work}
This work investigated whether semantic door cues can improve predictive geometric occupancy mapping in indoor environments containing door-induced ambiguity. By comparing a geometry-only control model with a semantic-cued model trained on the same modified dataset, we isolate the effect of adding explicit door semantics to the predictive mapping input.

The quantitative results show that semantic cues have limited effect on full-map prediction, but substantially improve localized prediction around doors. 

Within the unknown-region cells inside a 10-pixel radius of annotated door locations, the semantic-cued model reduced L1 error from 0.004342 to 0.000025 and improved F1 and IoU from 0.031311 and 0.015905 to 1.000000 and 1.000000, respectively. 

Qualitative results further show that the semantic-cued model more consistently predicts connected structures beyond door regions, while the geometry-only model often terminates predictions at door boundaries.

These findings suggest that semantic information can improve predictive geometric reasoning in regions where occupancy geometry alone is ambiguous. Future work will extend this study to larger and more diverse datasets, richer semantic cues beyond doors, automatic semantic extraction using vision-language or semantic perception models, and downstream evaluation in exploration and navigation tasks.

\section*{Acknowledgment}

One illustrative figure in this paper was generated using OpenAI DALL·E to support conceptual visualization. The generated image was not used as experimental data or evaluation results. AI-based tools were also used for grammar refinement and text editing assistance during manuscript preparation.

\bibliographystyle{IEEEtran}
\bibliography{IEEEabrv,mybib}

\end{document}